%% file: main.tex
\documentclass{ieeetj}
\usepackage{cite}
\usepackage{amsmath,amssymb,amsfonts}
\usepackage{algorithmic}
\usepackage{graphicx,color}
\usepackage{textcomp}
\usepackage{xcolor}
\usepackage{hyperref}
\usepackage{algorithm,algorithmic}
\def\BibTeX{{\rm B\kern-.05em{\sc i\kern-.025em b}\kern-.08em
    T\kern-.1667em\lower.7ex\hbox{E}\kern-.125emX}}
\AtBeginDocument{\definecolor{tmlcncolor}{cmyk}{0.93,0.59,0.15,0.02}\definecolor{NavyBlue}{RGB}{0,86,125}}

\newcommand{\LELR}[0]{\ifx\blind\undefined LELR\else The rover\fi}

\newcommand{\lelr}[0]{\ifx\blind\undefined LELR\else the rover\fi}

\def\OJlogo{\vspace{-4pt}$<$Society logo(s) and publication title will appear here.$>$}
\def\seclogo{\vspace{10pt}$<$Society logo(s) and publication title will appear here.$>$}

\def\authorrefmark#1{\ensuremath{^{\textbf{#1}}}}

\begin{document}
\receiveddate{21 February, 2025}
\reviseddate{19 December, 2025}
\accepteddate{5 January, 2026}
\doiinfo{10.1109/TFR.2026.3652156}

\markboth{}{\ifx\blind\undefined
Krawciw {et al.}
\else
Anonymous
\fi}

\title{Lunar Rover Cargo Transport: Mission Concept and Field Test}

\author{
\ifx\blind\undefined
Alexander Krawciw\authorrefmark{1}, Graduate Student Member, IEEE, Nicolas Olmedo\authorrefmark{2},\\ Faizan Rehmatullah\authorrefmark{2}, Maxime 
Desjardins-Goulet\authorrefmark{3}, Pascal Toupin\authorrefmark{3}, \\and Timothy D. Barfoot\authorrefmark{1}, Fellow, IEEE
\else
Authors Omitted for Anonymous Review
\fi}
\ifx\blind\undefined
\affil{Institute for Aerospace Studies, University of Toronto, Toronto, ON, Canada}
\affil{MDA Space, Brampton, ON, Canada}
\affil{Centre de Technologies Avancées, Sherbrooke, QC, Canada}
\corresp{Corresponding author: Alexander Krawciw (email: alec.krawciw@mail.utoronto.ca).}
\authornote{This work was made possible through the Canadian Space Agency's Lunar Surface Exploration Initiative and an NSERC Alliance Grant. \\A. Krawciw was supported by a Vanier Canada Graduate Scholarship. }
\fi

\begin{abstract}
In future operations on the lunar surface, automated vehicles will be required to transport cargo between known locations. Such vehicles must be able to navigate precisely in safe regions to avoid natural hazards, human-constructed infrastructure, and dangerous dark shadows. Rovers must be able to park their cargo autonomously within a small tolerance to achieve a successful pickup and delivery. In this field test, Lidar Teach and Repeat provides an ideal autonomy solution for transporting cargo in this way. A one-tonne path-to-flight rover was driven in a semi-autonomous remote-control mode to create a network of safe paths. Once the route was taught, the rover immediately repeated the entire network of paths autonomously while carrying cargo. The closed-loop performance is accurate enough to align the vehicle to the cargo and pick it up. This field report describes a two-week deployment at the Canadian Space Agency’s Analogue Terrain, culminating in a simulated lunar operation to evaluate the system's capabilities. Successful cargo collection and delivery were demonstrated in harsh environmental conditions. 
\end{abstract}


\maketitle

\section{INTRODUCTION}
\IEEEPARstart{T}{he} Artemis program is NASA's plan to return astronauts to the Moon \cite{Artemis}. This program has already completed the successful launch of the Orion spacecraft in 2022\cite{Artemis1}  and is working towards astronauts on the lunar surface by 2027 \cite{Artemis3}.
Beyond an initial touch-down, there are long-term plans to create a lunar outpost on the moon's surface.
Outpost structures will house astronauts and support science on the Moon. 
A key challenge to supporting such an operation is safely transporting building materials and supplies to assemble and maintain the operation. 
Uncrewed landers would make deliveries away from the settlement for safety reasons. 

\begin{figure}
    \centering
    \includegraphics[width=1.0\linewidth]{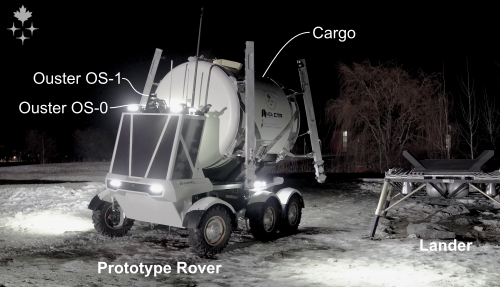}
    \caption{The Lunar Exploration Light Rover carrying cargo as it drives away from the lander on the right. The cargo transport mission is carried out under simulated lunar south pole lighting conditions and the ground is illuminated by the vehicle's headlights and the ``sun gun.'' Two Ouster lidars are used for mapping and hazard detection. (Image Credit: CSA) }
    \label{fig:landerVehicle}
\end{figure}
This creates a lunar logistics problem: large pieces of cargo need to be reliably transported from the lander sites back to the habitat and empty cargo must be removed \cite{mobility2024}. 
The long distances involved, currently estimated at 5 km \cite{cargo2024}, motivate using an autonomous cargo vehicle, allowing operations to occur without astronauts on the Moon, or freeing them to perform research and science instead of cargo transport. 

This paper presents a mission concept and associated field test of a lunar logistics mission carrying heavy cargo in a precise manner that enables future operations in the challenging lunar context. 
This field report takes existing lidar teach and repeat pipelines and evaluates in them in a new application domain to evaluate suitability and remaining challenges for future development. 
This is preparatory
\ifx\blind\undefined
 Lunar Surface Exploration Initiative
\fi work to develop and test technologies and concepts of operation that could enable future \ifx\blind\undefined Canadian \fi robotic contributions at the lunar surface.

\subsection{MISSION CONCEPT}
To support cargo transport operations on the Moon, we assume that there exists a known set of landing sites, a known set of cargo delivery locations, and a long distance of relatively unknown terrain between them. 
Satellite imagery and depth maps can be used in strategic mission planning, but will not contain enough information to allow precise path planning of a rover. 
In addition, we propose having a dedicated system at the landing site that allows the cargo to be loaded using a crane from a lander onto a transport vehicle. 
Once the cargo has been loaded, it will be self-sufficient. 
The rover's requirement is to park precisely relative to a location of interest; the cargo will be able to lift itself off the vehicle to perform the final docking procedures. 

These landing sites will be used many times, which means that there is an opportunity to leverage previous driving experiences between the lander and the habitat to speed up autonomous navigation. 
The teach and repeat framework \cite{Furgale2010} is ideally suited for this type of mission. 
The first time the rover navigates a route, it maps the path in a \textit{teach pass}.
After just one drive, the rover can autonomously repeat it in either direction along that route. 
In this concept, the first time the rover travels between the lander, habitat, and any other locations of interest, a network of paths is taught by using short, remotely specified path segments.
Another benefit of this exploration approach is the ability to undo if the progress is unsatisfactory. 
A small section of the recent path can be reversed, and then a new teaching branch can begin in a new direction \cite{stenning_planning_2013}.

In subsequent missions, the network of paths can be reused to repeat from the rover's current location to the location of interest. 
The critical benefit of a teach-and-repeat approach is autonomous traversals occur in the same path each time.
This means that if the path drives around a boulder or hazard, it will continue to do this every time without having to perform terrain assessment to observe these obstacles again if the environment is static. 
This reduces computational requirements and allows for more continuous drives between locations of interest, improving energy efficiency and saving time.
The only dynamic hazards on the moon will occur in known regions, so operators can move between operational modes depending on the environment.

In this concept, we assume that the cargo is an automated system itself. 
The cargo connects to the rover to allow the cargo to access rover data and to provide diagnostics back to remote operators through the rover's existing communication channels. 
Smart cargo provides long-term mission flexibility as new sensors that may be required for docking can be included at the cargo's time of launch rather than upfront with the vehicle. 

\subsection{FIELD TEST}
To evaluate this mission concept, \ifx\blind\undefined
 the Canadian Space Agency's (CSA) Lunar Exploration Light Rover (LELR) \cite{LELR} \else a one-tonne-scale rover \fi was upgraded with new avionics and outfitted with lidar for situational awareness and compatibility with Lidar Teach and Repeat \cite{Qiao2024, Burnett2022}.
Additionally, an actuated cargo was designed to be loaded onto the flat rear bed of the vehicle and carried into position.
\autoref{fig:landerVehicle} shows \lelr \ as it drives away from the lander towards the habitat. 
The simulated mission executes a small-scale version of the mission concept defined in Section~\ref{sec:mission}.

To make the field test as representative of a lunar deployment as possible, the testing was carried out on the Analogue Terrain of the CSA \cite{CSAMarsYard}.
The Analogue Terrain is 120 m x 60 m and contains sand, large rocks, and larger hill features that challenge rover drive systems and mimic the surface terrains of Mars. 
The operations were conducted remotely, except for a safety officer who maintained line of sight with the vehicle but was not part of the operational team. 
The remote operations were bandwidth limited to 2 Mbps and a delay of five seconds was inserted in both directions: sending requests to the rover and receiving data back from it. 
Finally, to recreate the long shadows and hard illumination caused by the sun at the lunar south pole, an 18 kW spotlight, called the sun gun, illuminated the otherwise dark terrain. 
Some tests were performed under sun-gun illumination and others in complete darkness.

\section{BACKGROUND}
Cargo transport will be a paradigm shift for the autonomy of extraterrestrial rovers. 
Existing rovers such as the Mars Exploration Rovers (Spirit and Opportunity) \cite{maimone_autonomous_2006}, Curiosity \cite{Rankin2019}, and Perseverance \cite{verma_autonomous_2023} only visit locations once. 
In missions like these, exploration is the focus. 
In practice, most missions to date rely on a high level of remote operator involvement for motion planning and hazard avoidance \cite{connell_ground_2023}. 
To explore, rovers need to perceive nearby hazards, plan around them, and estimate their current localized position to achieve strategic mission objectives \cite{wong_adaptive_2017}. 
In our cargo transport mission concept, exploration capabilities will be required to complete the first drive to new landing sites or habitat delivery locations. 

In the static lunar environment, hazard avoidance is performed primarily by assessing the geometric extent of the rover from stereo depth reconstruction \cite{gerdes_hazard_2020} and additionally considering the suspension of the vehicle as it moves over three-dimensional terrain \cite{otsu_clearance_2020}.
Unmanned ground vehicles on Earth often use lidar for terrain assessment off-road \cite{krusi_driving_2017, shan_lidar_2024} but so far no lidar has been sent on an extra-terrestrial rover mission \cite{cauligi_shadownav_2023}.

Strategic planning for exploration is based on the scientific objectives of the mission. 
Short-range path planning is used to plan short routes that deviate from the strategic plan in the presence of obstacles \cite{sutoh_right_2015}. 
Considerations such as slip estimation \cite{inotsume_slip_2020}, communication reliability \cite{verma_autonomous_2023}, and illumination of solar panels \cite{otten_strategic_2018} can all be included to find safe and efficient routes. 
Variations of sampling-based motion planners \cite{orthey_sampling_2024} are often used to select obstacle-free routes and optimal routes are chosen based on minimizing the energy consumed and the rover-specific considerations above.

Low-drift odometry remains important to cargo vehicle navigation and the ability to use maps generated in the operational environment opens new avenues for algorithm development and testing. 
Due to its high reliability, visual stereo odometry is the primary method of relative state estimation used on extraterrestrial rovers \cite{maimone_vo_2007, verma_enabling_2024}. 
However, if operating in illuminated regions near the lunar south pole, the resulting shadows will cause significant challenges for visual methods because stereo camera mapping quality degrades in shadows \cite{gerdes_hazard_2020}.
ShadowNav \cite{cauligi_shadownav_2023} uses a Hapke model of light from an illumination source on the ego-vehicle. 
This allows the vehicle to localize against known craters in a depth elevation map during lunar night. 
Fang et al. \cite{fang_ray_2021} use holes in stereo to find shadows and use ray tracing to help determine the relative orientation of the sun to the vehicle exploiting the presence of shadows as a feature that can be used to improve localization. 
Lidar odometry methods \cite{dellenbach_ct-icp_2022, xu_fast-lio2_2022} consistently outperform stereo odometry on standard on-road driving benchmarks \cite{agostinho_vo_2022}.
In addition to low drift, lidar methods are invariant to the ambient illuminations and will localize as shadows change. 
Until a 3D flight lidar is validated, it is difficult to assess the impact of the solar interference with a lidar at the lunar south pole.

Other field tests have been performed to evaluate cargo transport for a lunar environment.
The ATHLETE robot \cite{wilcox_athlete_2011} was designed as a rover that could be packaged with the cargo. 
It features six actuated arms with small wheels at the end of each. 
Cargo is carried on a central hexagonal platform. On Earth, ATHLETE could carry up to 300 kg of payload \cite{wilcox_athlete_2009}.
Large cargo cylinders mocking habitat modules were joined into a larger habitat complex in a large-scale field test. 
The unique actuated arm design enables ATHLETE to move in redundant ways. 
Under normal operation, the platform rolls on its six wheels and steers holonomically by twisting the arm. 
However, in case the vehicle gets stuck because of a large rock or soft sand, it can walk by lifting its legs up and down. 
This minimizes robot mass because smaller wheels and motors can be used.
An extension of ATHLETE was proposed to collect and receive payloads. 
The Tri-ATHLETE concept splits an ATHLETE robot into two half robots with three arm-legs each. 
The two vehicles approach a cargo from either side and latch onto it. 
The pallet of the cargo forms the central platform and rigidly links the two Tri-ATHLETEs into one functioning system. 
Although the half-rovers are less stable, they can still safely navigate without payloads. 

Other, more conventional rover concepts have been proposed for cargo missions. 
In their full-scale lunar mission analysis, Akin et al. \cite{akin_lunar_2025} propose using a rover with a U-shaped chassis that can align around standard cargo pallets. 
In addition to the large transport capabilities of a large rover, they suggest that certain payloads could be attached to a fleet of similar rovers to allow for regolith clearing and site preparation before habitat construction. 
A fleet of smaller multi-purpose rovers would be the only way to achieve the required scale or regolith clearing with planned launch capabilities.

For a cargo mission, the rover will transport loads between known places repeatedly.
This motivates the inclusion of maps created on the robot for continued operation in the future. 
VT\&R uses visual odometry to build up a topometric pose graph that allows relative localization of the robot during autonomous operation. 
The teach and repeat paradigm is simple: while exploring, capture features from odometry and store them in a local submap.
The submaps are joined together topologically with the relative transformations estimated from odometry. 
These two steps already occur to explore using odometry, so storing the relevant data structure is a lightweight addition to the system. 
Storing the relative transformations of the robot also encodes the robot's path. 
This process of odometry and mapping is the teach phase and the path the robot drives is called the teach path. 
Once the robot has driven the teach path, it can immediately drive it again autonomously. 
Any autonomous traversal is a repeat of the teach path. 
While repeating, the current sensor input is compared to the map to localize the robot with respect to the teach path. 
The relative position of the rover to the path forms the error term for a controller that brings the robot back to the teach path. 
Using only relative information enables robots to repeat the path accurately, even if the odometry used to teach it drifted.
In another lunar-simulated field test, VT\&R was evaluated on Devon Island, in the high arctic \cite{Furgale2010}.
Over 32 km of driving was completed with over 99\% autonomy rate.
A challenge with stereo camera teach and repeat is the dependence on ambient lighting. 
Features captured during the day look different from those captured at night under headlight illumination. 
Extensions of VT\&R are robust to illumination changes either by repeating routes while the illumination changes to capture multiple experiences \cite{paton_bridging_2016} or by using deep learning to extract feature descriptors that can be matched directly between day and night \cite{Chen2023}.  
Switching to lidar as the primary sensor \cite{Burnett2022} prevents issues related to ambient illumination.
\ifx\blind\undefined
This work is an application of the lidar odometry and localization developed by Burnett et al. \cite{Burnett2022} and the model-predictive control developed by Qiao et al. \cite{Qiao2024}.
\fi

\ifx\blind\undefined
The Lunar Light Exploration Rover was initially designed to support uncrewed exploration or manual piloting from astronauts onboard \cite{mccoubrey_canadian_2012}.
Its significant payload capacity allowed flexibility in adding science instruments or human pilots.
Visual Teach and Repeat (VT\&R) \cite{Furgale2010} was demonstrated as an autonomy solution that allowed the rover to use stereo imagery to follow a mapped route. 
Using a combination of the visual odometry from VT\&R with local and global planning, LELR completed a large-scale test campaign in 2015 \cite{bakambu_maturing_2016}.
\fi

For the Artemis mission, a large volume of lunar cargo will be transported from landers to the Artemis base.
NASA estimates a cargo demand of up to 10,000 kg per year \cite{cargo2024}.
Currently, planned cargo includes habitation modules, communication systems, and the logistics infrastructure. 
In addition to one-time items required to establish operations, resupply missions bring food, water, and spare parts.
All of this cargo will require different surface mobility solutions to move goods to their appropriate final destination. 
High cargo volumes motivate the use of autonomous transport vehicles on the lunar surface.

\section{MISSION PLAN}
\label{sec:mission}
The field test is designed to evaluate the viability of the proposed mission concept and expose new technical challenges for further development. 
At a high level, there are three stages of the tests: a semi-autonomous stage where a teach is performed, an autonomous stage where the mission is completed in reverse, and a forward autonomous stage that repeats all of the mission motions completed in the teach.
In a lunar deployment, the manual control stage would only be performed once to complete the long traverse between the habitat and lander sites. 
This could be any mixture of teleoperation, astronaut control, or a slower autonomy stack.
Autonomous repeats can occur any number of times after the first teach.
Even if the landers change positions, a repeat can be used to bring a rover near the landing site and then another operating mode can perform the fine alignment with the lander.

\begin{figure}
    \centering
    \includegraphics[width=0.95\linewidth]{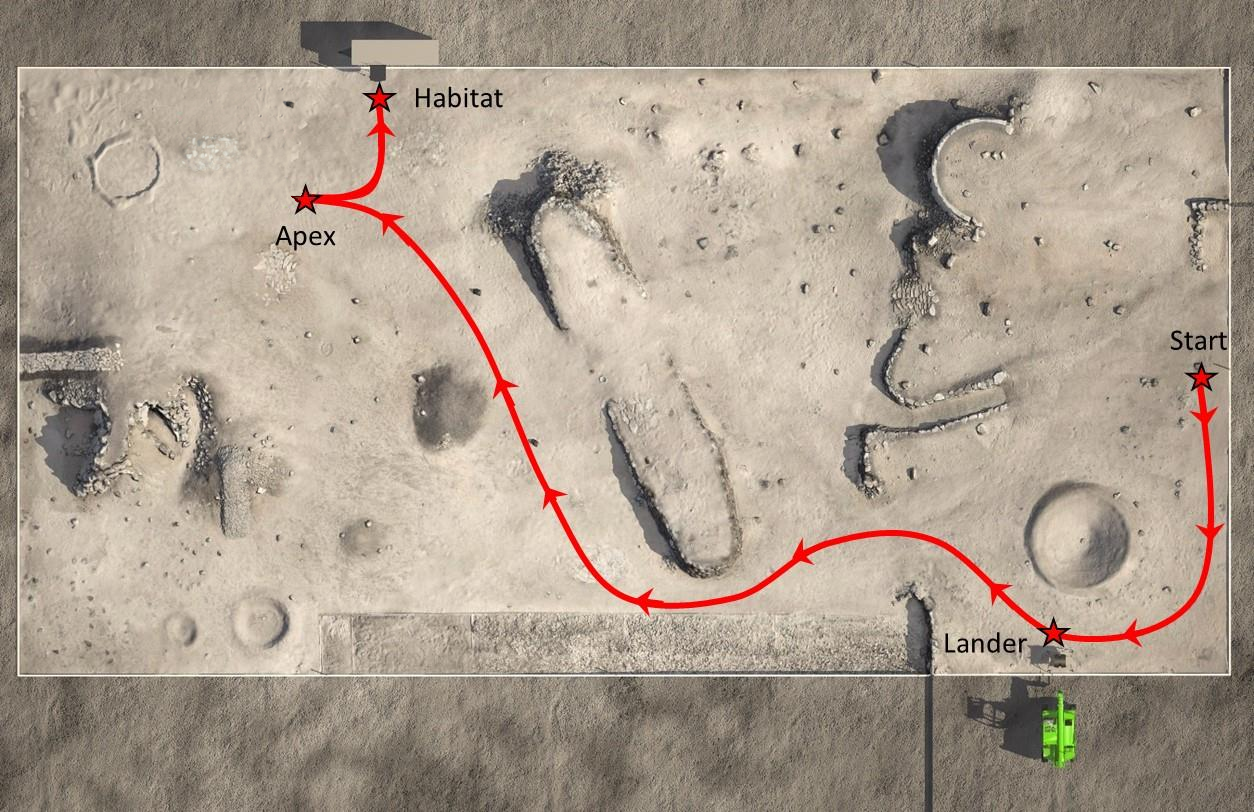}
    \caption{The key locations for the simulated lunar mission. The lander and habitat require precise relative alignment. The other locations are approximate and will be selected based on an operator's assessment of navigability. (Image Credit: CSA)}
    \label{fig:mission}
\end{figure}

\autoref{fig:mission} shows the high-level route that was selected from satellite and elevation images. 
The apex is a location close to the habitat where the rover will switch direction to align. 
The lander and habitat require the rover to align within a prescribed tolerance to ensure successful cargo operations. 
The semi-autonomous mission stage has seven steps: 
\begin{enumerate}
    \item Drive from the start to the lander semi-autonomously and align with the lander.
    \item Transfer cargo from the lander to the vehicle using a crane.
    \item Drive forward from the lander to the apex semi-autonomously.
    \item Drive backward from the apex to the habitat aligning the cargo.
    \item Unlatch and raise the cargo from the vehicle bed.
    \item Drive the vehicle forward so that it is clear of the cargo.
    \item Lower the cargo so that it is completely aligned.
\end{enumerate}

At this point, the rover has demonstrated a successful delivery. 
The second mission stage essentially resets the environment by performing the entire mission in reverse. 
The most precise motion is the second operation: reversing the vehicle back under the raised cargo. 
To latch successfully, the rover requires positioning of $\pm 7.5$ cm longitudinally and $\pm 10$ cm laterally.
Additionally, the cargo legs have $\pm 15$ cm of clearance from the centerline and there is a $20$ cm space between the vehicle's cab and the cargo. 
The mission reset stage has seven opposite steps:
\begin{enumerate}
    \item Raise the cargo so that the rover can park underneath.
    \item Reverse the vehicle under the cargo.
    \item Lower the cargo onto the vehicle bed and latch to it.
    \item Drive forward to the apex autonomously using a repeat.
    \item Drive in reverse from the apex to the lander using a repeat.
    \item Unlatch the cargo and move it from the vehicle to the lander using the crane.
    \item Drive in reverse from the lander to the start.
\end{enumerate}

After resetting the environment, the first mission stage is executed again with autonomous driving for all motions. 
The lander and habitat positions are recorded precisely by the teach path so operators simply instruct the vehicle to repeat to them. 
The autonomous mission has four steps:
\begin{enumerate}
    \item Drive forward to the lander autonomously using a repeat.
    \item Load the cargo using the crane.
    \item Drive forward to the apex and then reverse align to the habitat autonomously using repeat.
    \item Deploy the cargo.
\end{enumerate}

Autonomous missions should occur more quickly than semi-autonomous missions. 
To compare the two missions, the completion time, path consistency, and cargo handling were assessed. 

\begin{figure}[t]
    \centering
    \includegraphics[width=\linewidth]{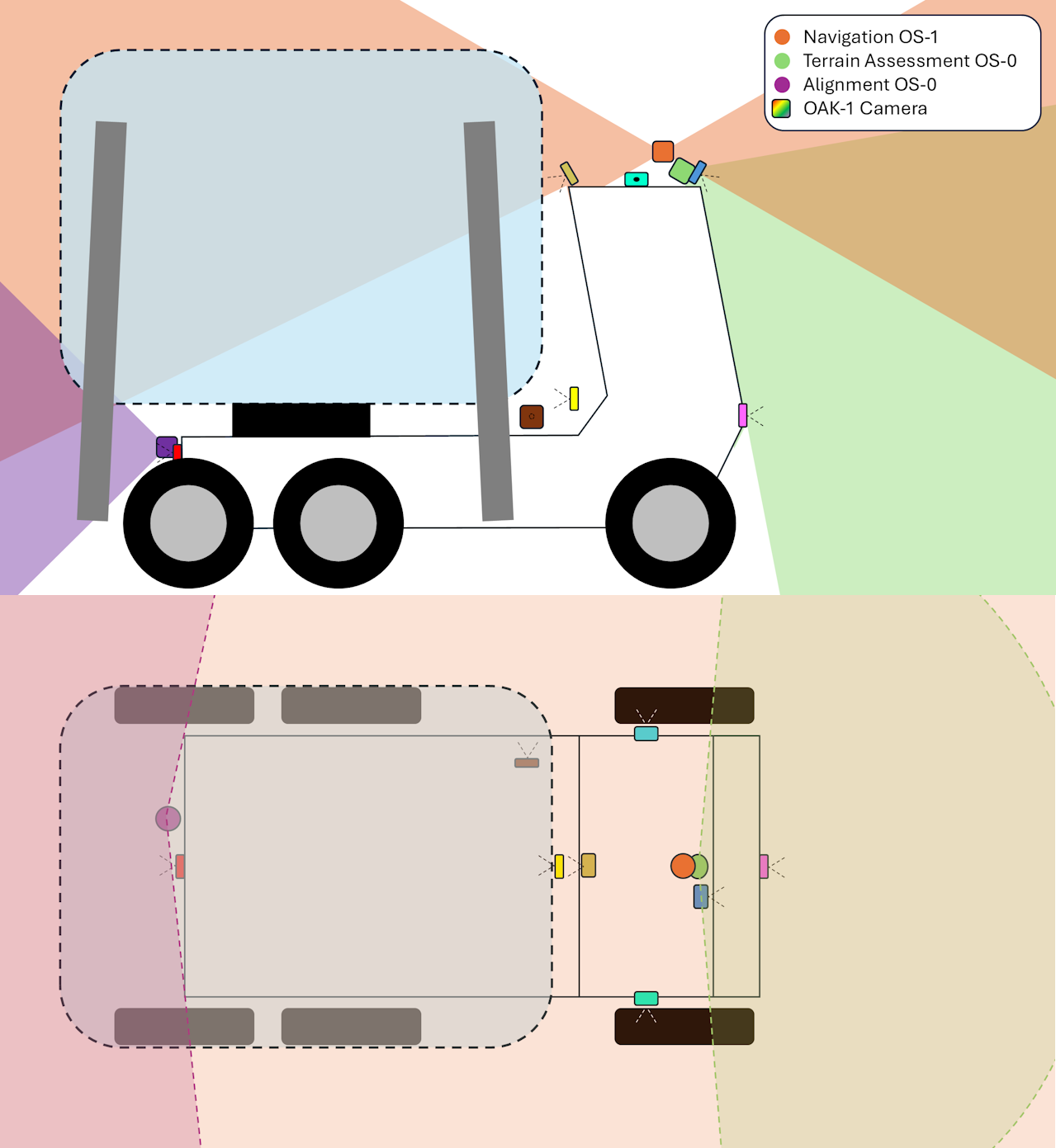}
    \caption{A side and top view showing the locations of the three lidar sensors and eight cameras. The lidar fields of view are denoted by the swept area filled with each sensor's colour. Note that the cargo (\textit{light blue}) reduces the navigation lidar's (\textit{orange}) field of view as the vehicle carries it. The direction of each camera is marked with a dashed triangle.}
    \label{fig:sensor_positions}
\end{figure}

\begin{figure}[t]
    \centering
    \includegraphics[width=\linewidth]{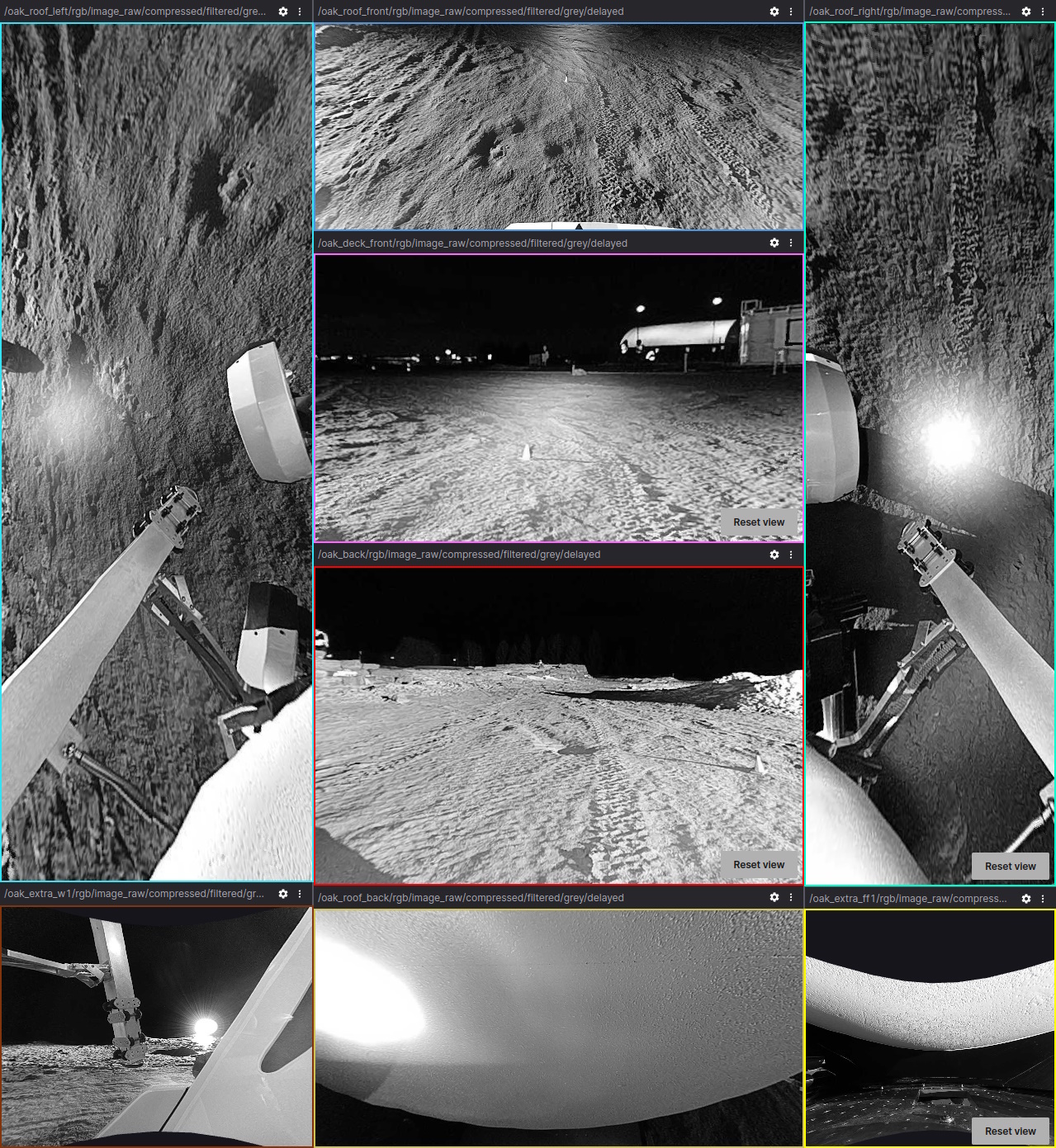}
    \caption{A composite view of all eight cameras mounted to \lelr. The border colours correspond to the camera's colour in \autoref{fig:sensor_positions}.}
    \label{fig:cams}
\end{figure}

\section{HARDWARE}
\LELR \ is a six-wheel drive vehicle with two independently steered wheels controlled by electric actuators and four fixed wheels. The steering can replicate an Ackermann geometry virtually. The rover possesses the capability for zero-radius turning through skid-steering, achieved by independently modulating the rotational velocities of the left and right wheel sets via dedicated electric wheel motors. This allows for differential wheel speeds, including counter-rotation, enabling highly manoeuvrable, on-the-spot pivoting. 
However, under typical operational conditions, the rover employs a steering mechanism analogous to that of a conventional automobile, prioritizing controlled, Ackermann-like steering dynamics to optimize stability, minimize tire wear, and reduce energy consumption on standard terrains.
The motion of the rover is controlled by a low-level dynamics controller that converts body velocity commands from high-level control loops into the torques for the six wheels and angles for the steering actuators \cite{chhabra_dynamical_2016}.

The rubber tires have a diameter of 0.763 m allowing obstacles up to 30 cm high to be cleared. \LELR \ is powered by a set of lithium-ion batteries with a total capacity of 12 kWh at 96V. The vehicle has an unloaded mass of 800 kg and can carry up to 300 kg at a maximum velocity of 13 km/h. 
The front cab was added to improve the viewpoint of the primary sensors, position lights, and emulate the radiators and solar panel arrays required for energy transfer on the Moon.

Remote communications used a 9 dBi antenna at 2.4 GHz. The link has an allowable range of approximately 300 m from the base station. 
The ground control station antenna is placed sufficiently far away from the GCS building to reduce interference.

\subsection{SENSORS}

\LELR \ has three lidars, eight cameras, an inertial navigation system (INS), and position feedback on the cargo latches.
The primary navigation sensor is an Ouster OS-1 128 lidar \cite{ouster_128} mounted parallel to the body and centred on top of the cab. 
This lidar has the largest field of view, which is beneficial while mapping. 
The rear third of the field of view is obstructed when the cargo is loaded, so the view of the environment changes depending on the cargo configuration.
An Ouster OS-0 128 is tilted towards the ground and mounted on the top of the cab. 
This lidar provides a view right in front of the vehicle, filling in the near-range blind spot of the OS-1 and providing terrain information for navigability assessment.
The third lidar is an Ouster OS-0 128 mounted below the cargo on the rear of the vehicle. 
The rear view is required to operate \lelr \ semi-autonomously while driving in reverse.
Self-occlusions obstruct the ground near the vehicle even when there is no cargo.
This lidar captures the habitat precisely as the vehicle reverses to align the cargo. 

The eight cameras improve the situational awareness of remote operators but are not currently integrated into the autonomy stack. 
The front and rear deck cameras show the terrain along the path of the rover. 
Four cameras are mounted on the cab roof and tilt down to show the footprint of the vehicle. 
Two additional cameras are mounted on the cargo deck for operational support. 
One camera views the cargo alignment guide to monitor cargo loading. 
The other looks sideways at the lander as the rover drives alongside it.
\autoref{fig:cams} shows a view from all eight cameras after the cargo has been loaded onto the vehicle.

A Novatel Span G370 inertial navigation system (INS) is mounted to the vehicle's roof and provides ground truth for evaluating path tracking. 
GPS postprocessing was used to improve the precision of the INS pose estimates with Inertial Xplorer software. 

Remote cargo handling requires sensor feedback to ensure that each step is completed safely and correctly. 
The actuators that clamp the cargo onto the vehicle's bed have position and current feedback. Position feedback is used when releasing the cargo to ensure that the hooks are clear of the latching rod. 
Current feedback is used when latching the cargo as a proxy for force feedback. 
The cargo is securely clamped when the actuator current exceeds a calibrated threshold. 

While the cargo is lifting, an IMU inside it provides feedback on roll and pitch.
This information is used to adjust the power for each leg and keep the cargo safe and level for the final docking. 

After the rover has driven clear of the cargo, the legs retract to achieve the target alignment height.
A downward waterproof ultrasonic distance sensor measures the height of the cargo off the ground.

\subsection{CARGO}
\begin{figure}
    \centering
    \includegraphics[width=0.75\linewidth]{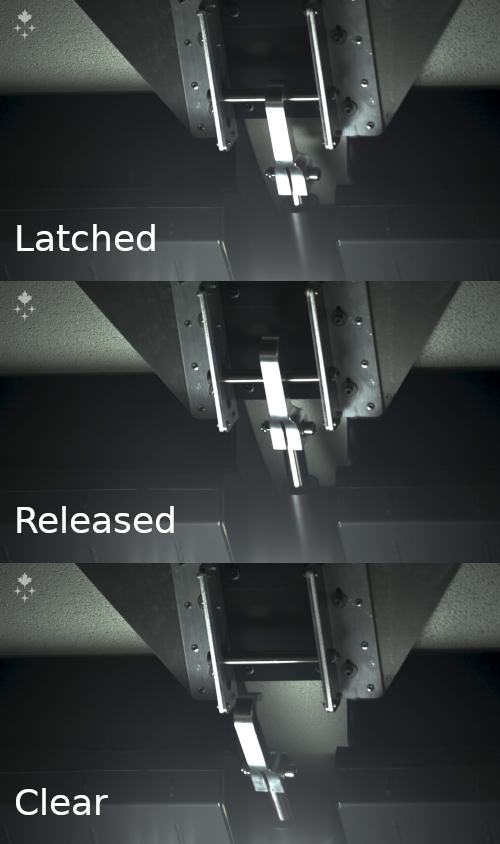}
    \caption{A three-frame sequence of a latch release. When the latch is fully engaged, as in the first pane, the cargo is secure and it is safe to drive the rover. When the latch is clear, loading and unloading operations can occur. (Image Credit: CSA)}
    \label{fig:latching}
    \vspace{-5mm}
\end{figure}

The cargo module measures 1.6 meters in diameter and 2.2 meters in length. It is constructed from expanded polystyrene which optimizes both structural integrity and weight efficiency. 
The interior of the cargo is hollow, providing accommodation for various electrical subsystems. 
The door frame facilitating access to these components was 3D printed to ensure precision and structural adequacy in a single, seamless piece.
The complete cargo assembly has a total mass of 182 kilograms, balancing the requirements for structural resilience, weight constraints, and operational functionality.

\subsubsection{Telescopic Leg Mechanism}

The cargo is supported by four independently actuated telescopic legs, each powered by electric actuators to allow precise deployment and retraction. 
A simplified plastic roller system facilitates smooth telescopic motion, with tolerances and openings designed to prevent small debris from obstructing movement.
Each leg allows 1.52 meters of travel, allowing it to retract within the rover's footprint.

When deploying the cargo, the legs begin fully retracted. 
The legs extend to elevate the cargo up to 0.5 m above the deck, allowing it to support itself independently of the rover.
With the cargo securely self-supported, the rover disengages and departs, leaving the cargo in its designated location.
After the vehicle has departed, the legs retract for final alignment with the habitat.

\subsubsection{Alignment and Load Distribution Systems}

The alignment guides feature 45$^\circ$ tapered surfaces, allowing for tolerance of misalignments up to ±150 mm between the rover deck and the cargo.
The contact surfaces of these guides are covered with 3.175 mm ultra-high-molecular-weight polyethylene sheets, which allows the cargo to slide into the correct position to be latched. 

The legs pivot with respect to the cargo body as they retract or extend. 
The leg end caps are designed to inhibit outward slipping while allowing controlled inward motion, improving the system's stability.

\section{AUTONOMY STACK}
The backbone of the autonomy used for this cargo mission is Lidar Teach and Repeat (LT\&R) \cite{Burnett2022}.
LT\&R has been shown to be effective for long-term navigation in harsh and unstructured environments such as forests, grassy plains, and deep snow \cite{Sehn2024, Qiao2024}. 
The method follows the same framework as Visual Teach and Repeat \cite{Furgale2010} but lidar offers a few advantages, especially for lunar deployments. 
Lidar sensors are invariant to ambient lighting. 
This means that even with the low elevation of the sun at the Moon's south pole, the sensor would not be blinded.
Long shadows that change as the Moon orbits make long-term feature matching more difficult for stereo camera localization \cite{allan_planetary_2019}. 

\begin{figure*}
    \centering
    \includegraphics[width=0.9\linewidth]{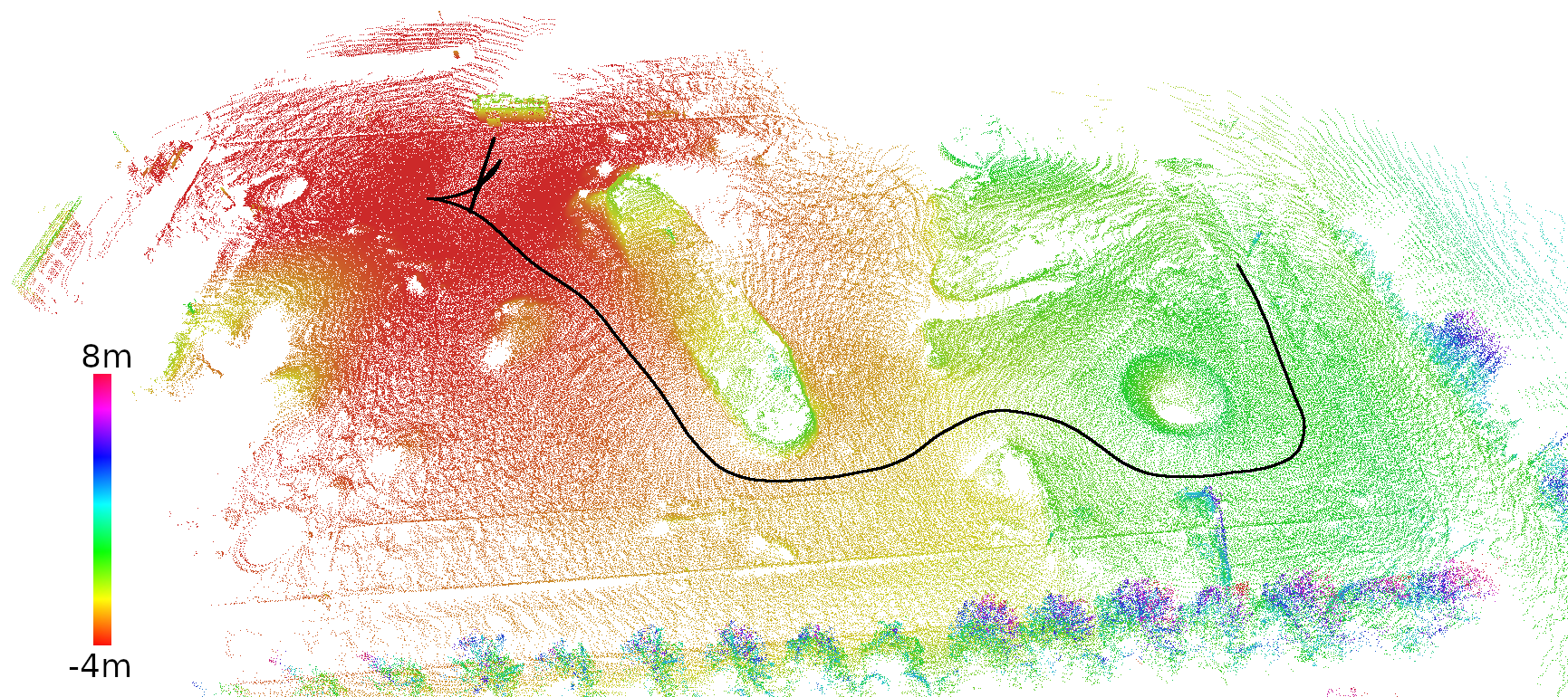}
    \caption{A lidar map of the CSA's Analogue Terrain created by accumulating the lidar scans from \lelr \ as it completed the demonstration in simulated lunar south pole lighting conditions. Points are coloured by their elevation relative to the start position. The rover's estimated path is shown in black.}
    \label{fig:odomMap}
\end{figure*}

The teach phase uses lidar odometry to estimate the relative transformations between nearby submaps along the route. 
The Ouster OS-1 lidar is used for odometry and mapping because it has the largest field of view around the vehicle. 
Our lidar odometry uses a robust variant of the Iterative Closest Point (ICP) algorithm \cite{Burnett2022}.
Point-to-point associations form the loss function but the cost weighting is based on the estimated normal of the map providing many of the optimization benefits of a point-to-plane approach. 
Adding robust cost functions rejects outlier points that do not match.  
This improves localization performance with scans that include the cargo matching to maps that do not contain it.  
The odometry drift was evaluated as 0.54\% on the Boreas dataset \cite{Burnett2022} using the Kitti metric, which is lower than visual odometry \cite{geiger2012we}.
The odometry drift is small enough to accumulate useful maps while operating. 
\autoref{fig:odomMap} is an accumulation of one scan every 3 m along the route, transformed using the odometry roll-out to that location. 
Qualitatively, the edge of the habitat (top middle) and other distinct features like craters are well defined.

While teaching new routes, a pose is stored every 30 cm or 10 degrees of motion.
Submaps are stored less frequently. 
The submaps are accumulations of recent scans, cropped to 40 m around the rover and voxel downsampled.
Accumulating scans means that blind spots can be filled and map points are distributed more evenly. 
Uniform map density improves localization if a different lidar configuration is used.
The operator requests branch and merge points to create an arbitrary network of paths.
During a repeat, the vehicle may take any route through the branches regardless of the order in which they were taught. 

The repeat phase allows the robot to autonomously navigate the route after it was driven.
Lidar odometry is computed first, and the compounded transformation of odometry and the prior localization estimate is used as a prior in the localization optimization.
When driving autonomously, the submap nearest to the robot is selected and ICP is used to estimate the robot's position within that submap. 
As the robot moves along the route, new submaps for localization are selected. 
The robust cost function in ICP improves localization: dynamic obstacles in either map or live scan are rejected as outliers.

The lidar odometry is accurate enough to use dead reckoning to close the control loop for distances a few meters long without an appreciable impact on the path tracking. 
While the full odometry and map localization pipeline usually runs in real-time (i.e. greater than 10 Hz), map localization may take longer depending on the vehicle's motion and the map's quality.
Instead of reducing localization quality to maintain speed, the exponential moving average of the runtime is used to decide on the fly whether to skip map localization. 
This ensures that the controller will receive state updates at 10 Hz and allows ICP map localization to take the appropriate number of iterations to converge.

Once localization is complete, the relative cross-tracking error between the vehicle and the taught path forms the input to a model-predictive-control (MPC) problem.
The teach path defines the target robot poses over a finite horizon. 
The MPC aims to follow the target path at an appropriate speed and uses a unicycle model centred between the four rear wheels to determine an optimal control command \cite{Qiao2024}. 
A speed scheduler adjusts the target forward velocity based on the curvature of the path. 
The command is a planar twist: forward speed and angular velocity.
The MPC solves for commands at 15 time steps spaced 0.25 s apart.
The first command from the solution is sent to the vehicle. 
With the 0.1 s time for each lidar scan, an updated MPC problem will be solved before the prediction window is complete. 
A low-level controller in the vehicle converts these twist commands into motor torques for all six wheels and a steering angle for the front two \cite{chhabra_dynamical_2016}.
To reduce the MPC's modelling error, a first-order lag is added to the angular velocity command.

\section{SEMI-AUTONOMOUS DRIVING}
Direct teleoperation of \lelr \ was deemed infeasible due to the limits placed on sensor bandwidth and the ten-second round-trip delay between the ground control station and the vehicle. 
A semi-autonomous driving paradigm was implemented to provide a mode of exploration based on human terrain assessment.
An operation begins with the approximate global location of the vehicle, based on lidar odometry estimates and other large features visible in satellite data. 
Sensor data is triggered by the driver and sent back to the ground control station. 
Depending on the direction of motion, the appropriate cameras and lidar are queried. 
After receiving the scans, the rover driver creates a short path that is safe for the rover to traverse.
To define the path, the driver uses a joystick to move a 3D model of the vehicle through the 3D point cloud map accumulated by the lidar. 
The model contains the same kinematic constraints as the real vehicle ensuring that the selected path is valid and smooth. 
If the operator is unsatisfied with the path rollout, they can remove part of it and try again or restart completely. 
Some parts of the mission are best defined using delta pose commands rather than virtual driving. 
For example, when driving out from underneath the raised cargo, the rover must drive straight forward to prevent a collision with the cargo legs. 
In this case, a change in the longitudinal position of the rover can be specified using a delta position.

\begin{figure}
    \centering
    \includegraphics[width=\linewidth]{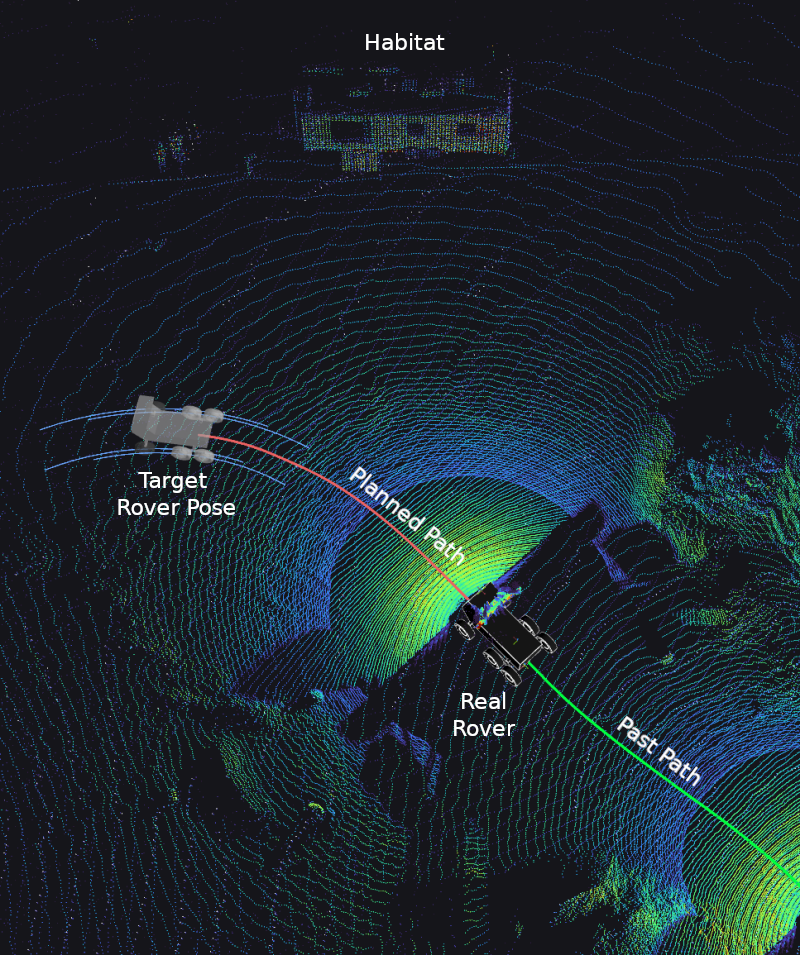}
    \caption{The final target rover pose (translucent grey) at the end of the semi-autonomous path segment (red). The red line denotes the target path remaining and the green line tracks the actual motion of the real rover (black). The point cloud map is an accumulation of scans stitched together using lidar odometry. The real rover will stop moving at the pose of the target rover. The blue lines show the current arc of the front and rear wheels based on the current joystick steering angle.}
    \label{fig:phantomRover}
\end{figure}

The paths always begin from the rover's current estimate of its location, which prevents aggressive path recovery behaviours at the beginning of motions.
Once the driver is satisfied with the visualization of the future robot motion, shown in \autoref{fig:phantomRover}, the rover tracks the path based on the rollout of lidar odometry measurements. 
The low drift rate of lidar odometry leads to accurate motions that reflect the desired paths. 

\begin{figure*}
    \centering
    \def\svgwidth{\linewidth}
    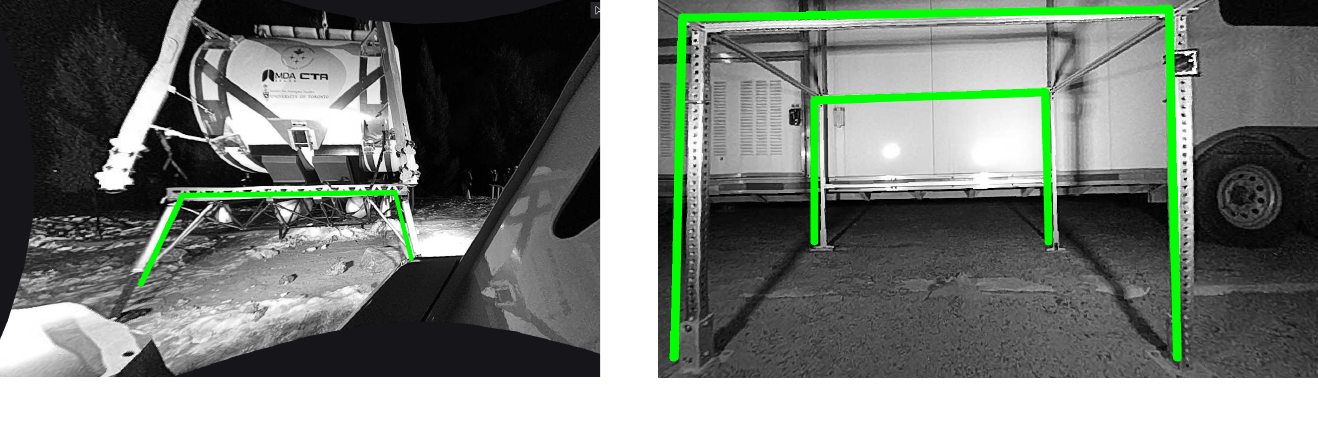
    \vspace{-5mm}
    \caption{The alignment guides overlaid with the static structures on the lander (a) and habitat (b). When the green lines cover the legs of the structures, operators know that the vehicle is aligned.}
    \label{fig:alignGuides}
\end{figure*}
When the rover approaches a location where precise alignment is required, for example, the habitat, relative measurements between the rover and the target are used to guide the next motions.
At the lander, the vehicle must align parallel to the cargo for a successful loading operation. The side deck camera has an alignment guide overlay to determine when the vehicle is in position. 
At the habitat, the rear lidar determines the distance between the vehicle and the docking port. 
The rear camera also has alignment overlays on the legs of the habitat. 
Alignment guides can be created based on known camera parameters and 3D models of the structures of interest.
The final manual alignment of the vehicle is shown at both the lander and habitat in \autoref{fig:alignGuides} a) and b) respectively.

\section{EXPERIMENTS}
The field test evaluation was performed after sunset from 5:45 p.m. to 10:30 p.m.
The temperature started at -3$^\circ$C at 6 p.m. and fell to -6$^\circ$C at 11 p.m. \cite{YHUWeather}.
More significantly, the wind speed was 46 km / h at 6 p.m. and gradually dropped to 24 km / h at 11 p.m. \cite{YHUWeather}. 
Due to strong winds, it was unsafe to lift the cargo with the crane three times as originally proposed in the mission plan. 
The cargo lift was performed only once during the semi-autonomous mission stage. 

\begin{figure*}
    \centering
    \def\svgwidth{\linewidth}
    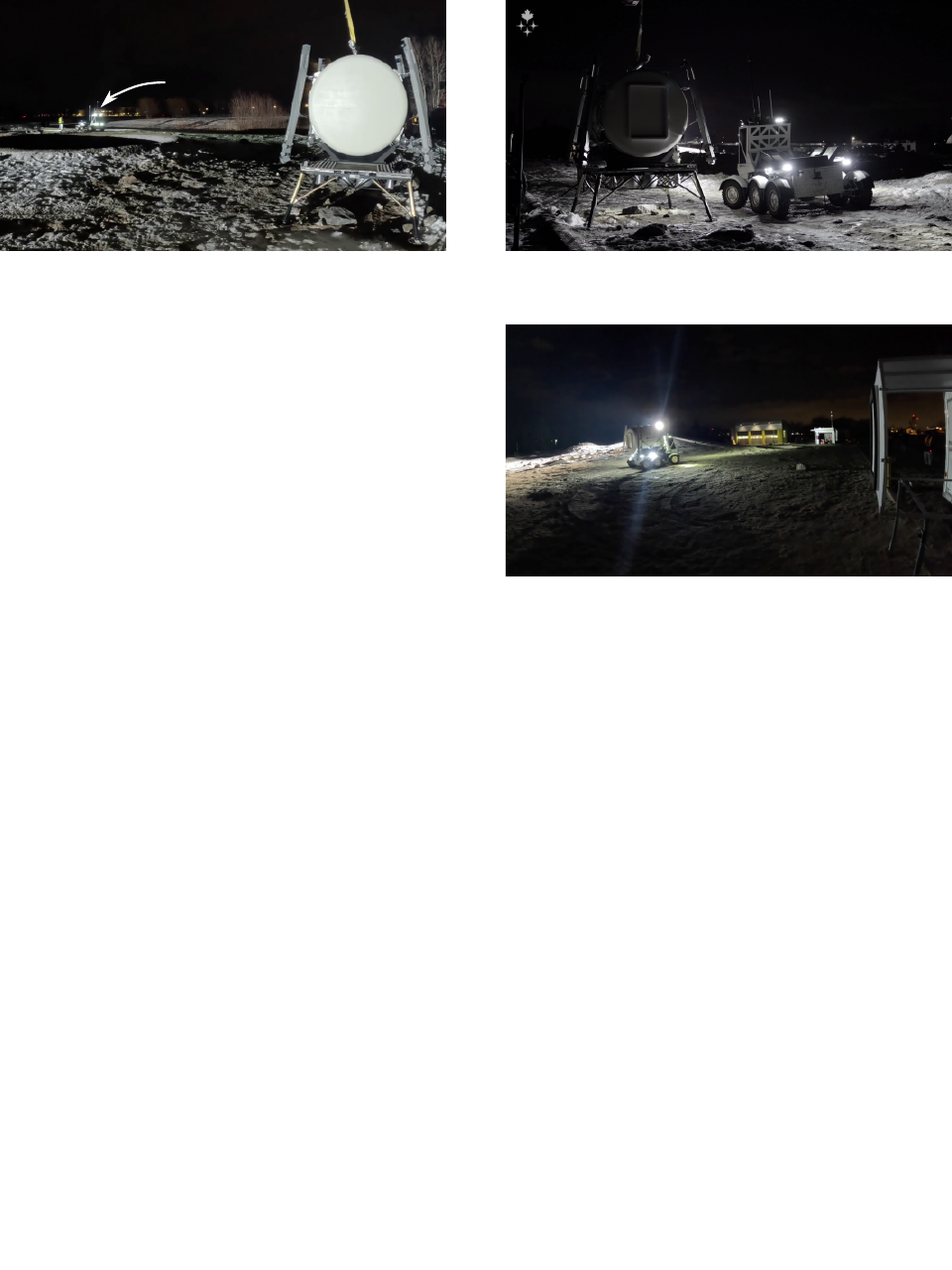
    \vspace{-5mm}

    \caption{A composite panel showing the snapshots of the rover as it completes the semi-autonomous mission plan. Image Credit: CSA}
    \label{fig:missionSteps}
\end{figure*}

\begin{figure*}
    \centering
    \includegraphics[width=\linewidth]{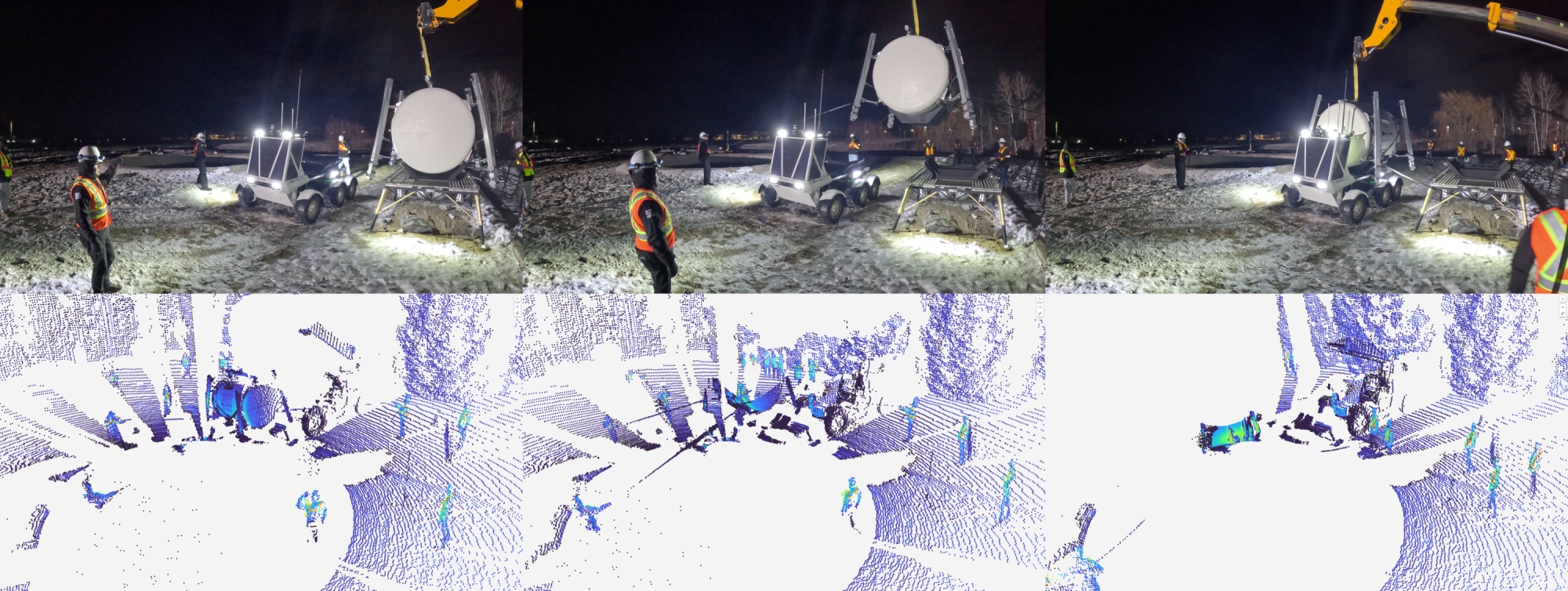}
    \caption{The cargo is lifted using the telehandler and loaded onto \lelr. The top row is a static camera, and the lower row is the synchronized view from the rover's OS-1 lidar. The high winds required the use of extra personnel with tow lines.}
    \label{fig:cargoLift}
\end{figure*}

Figures \ref{fig:missionSteps}a-h show the semi-autonomous mission stage as a sequence of frames. 
Initially, the rover is empty at the start (\ref{fig:missionSteps}(a)) and drives until it is aligned with the lander (\ref{fig:missionSteps}(b)). 
The cargo was loaded using the telehandler and supported with four human-guided tow lines.
In \autoref{fig:cargoLift}, the crane operation is performed. 
The rover can track the operation using the 360$^\circ$ view from the OS-1 lidar.
The lidar and rear deck camera allowed operators to measure precisely that the cargo was seated correctly and ready to be latched. 
The high winds necessitated four tow lines to prevent the cargo from swinging. 
Even in earlier testing with lower wind speeds, the cargo was difficult to carry with the telehandler. 
Once the cargo is loaded onto the vehicle bed, it is secured with a pair of latches.
The hook closes into the cargo and pulls down until it is tight. 
In \autoref{fig:latching}, the unlatching sequence is shown. 
With the cargo loaded, the rover drives the bulk of the route (\autoref{fig:missionSteps}(c)) and switches directions at the apex (\autoref{fig:missionSteps}(d)).
The alignment into the habitat occurs with \lelr \ driving in reverse. 
\autoref{fig:missionSteps}(e) shows the cargo during the final approach. 
Once the cargo is aligned, it can be released. 
First, the cargo is unlatched, as shown in \autoref{fig:latching}.
Next, the legs extend (\autoref{fig:missionSteps}(f)) until all four are touching the ground. 
As they extend, the cargo rises above the cargo deck until the state of \autoref{fig:missionSteps}(g).
Once clear, \lelr \ drives forward to leave the cargo behind (\autoref{fig:missionSteps}(h)).
The final height adjustment of the cargo occurs and the legs retract again to set the cargo to the desired final position. 

Three types of performance were evaluated: alignment accuracy with the lander and habitat, path-tracking error, and mission time. 


\begin{figure}
    \centering
    \includegraphics[width=\linewidth]{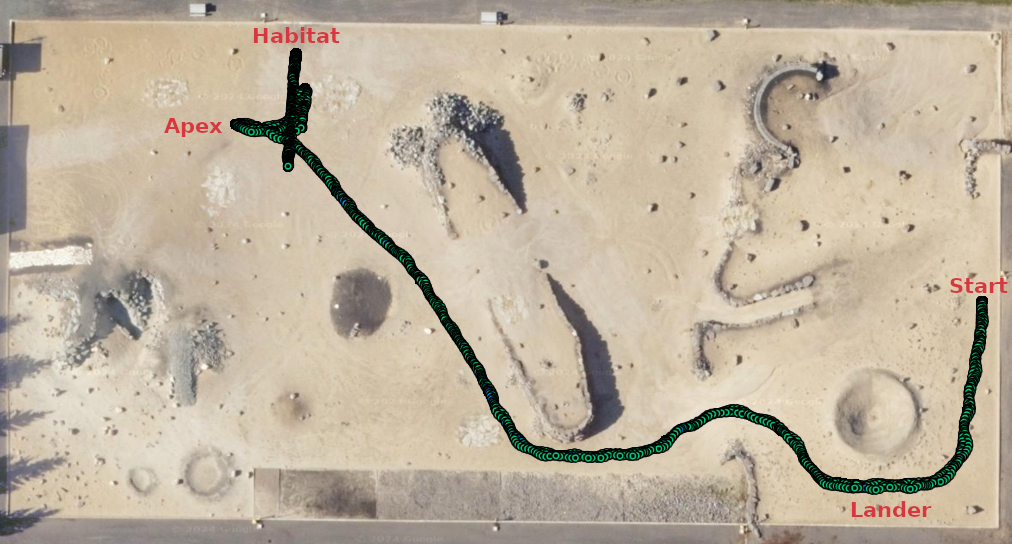}
    \caption{The ground-truth route of the rover superimposed on the satellite view of the Analogue Terrain. }
    \label{fig:gpsRoute}
\end{figure}

The rover's alignment quality is presented in \autoref{tab:alignments}.
The distance measurements were taken with a tape measure by the safety officer once the flight team declared that an operation was complete. 
When aligned with the lander, the rover should be parallel and between 1.5 - 2.0 m laterally away. 
Because the telehandler moves the cargo laterally from the lander to the rover, there is a wider margin for error. 
The heading should be within $\pm5^\circ$ of parallel.
The front of \lelr \ was 1.8 m and the rear was 1.7 m, both within the distance tolerance. 
The heading misalignment was 2.7$^\circ$ which is sufficiently parallel for the alignment guides to work properly. 

The alignment with the habitat targeted less than 0.15 m gap between the cargo and the habitat without impact. 
Laterally, the requirement is $\pm 0.05$ m for the centreline of the cargo with the habitat. 
The final position of the cargo was 0.17 m away from the habitat and 0.01 m laterally offset. 
The longitudinal gap was slightly too long, but the cargo was laterally aligned well. 

\begin{table}[]
    \centering
    
    \caption{Alignment results for the rover during the semi-autonomous teach mission phase.}
    \begin{tabular}{c|c|c}
        \textbf{Alignment Target} & \textbf{Value} & \textbf{Result} \\
        \hline
        Distance from Lander & 1.5 m - 2.0 m & 1.75 m \\
        Parallel Angle to Lander & 0 $\pm$ 5$^\circ$ & 2.7$^\circ$ \\
        Longitudinal Gap from Habitat & (0, 0.15] m & 0.17 m \\
        Lateral Misalignment from Habitat & 0 $\pm$ 0.05 m & 0.01 m
    \end{tabular}
    \label{tab:alignments}
\end{table}

\begin{table}[]
    \centering
    \caption{Path-tracking performance while performing the cargo mission. All values in meters.}
    \begin{tabular}{c|c|c|c}
        \textbf{Description} & \textbf{Mean $\pm$ $\sigma$} & \textbf{RMSE} & \textbf{Max Error} \\
        \hline
        LT\&R estimate at rear & 0.024 $\pm$ 0.112 & 0.101 & 0.401 \\
        LT\&R estimate at GPS & 0.015 $\pm$ 0.167 & 0.142 & 0.472 \\
        GPS measurement & -0.041 $\pm$ 0.248 & 0.252 & 0.573 \\
    \end{tabular}
    \label{tab:path_tracking}
\end{table}

\begin{figure}
    \centering
    \includegraphics[width=0.9\linewidth]{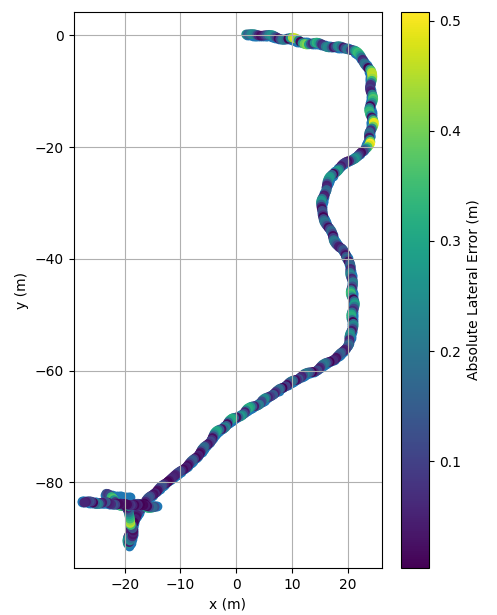}
    \caption{The absolute path-tracking error of LT\&R evaluated at the position of the GPS, near the front of the vehicle. The vehicle steers back and forth across the taught path. The largest errors occur near the sharpest turns. }
    \label{fig:ltrError}
\end{figure}

GPS was used to determine the ground truth of the rover's position.
\autoref{fig:gpsRoute} shows the route taken over the terrain.
The control point of the path-tracking controller is the center of the rear four wheels. 
The path-tracking errors are summarized in \autoref{tab:path_tracking}.
The LT\&R estimate at the rear is the controller error. 
This RMSE is 0.10 m.
Applying the known extrinsic calibration between the rear axle and the GPS generates the LT\&R estimates at the GPS. 
The RMS path-tracking error at the GPS is 0.142 m and the maximum is 0.472 m.
This error is visualized in \autoref{fig:ltrError}.
The Ackerman nature of this vehicle means that heading errors at the rear axle lead to larger lateral errors at the front of the vehicle. 
The GPS ground-truth path-tracking RMS error at the front of the rover is 0.252 m. The maximum error is 0.573 m.
This is larger than the LT\&R estimate in the same location suggesting that errors are a mixture of controller errors and localization errors.
Comparing the relative positions of the teach and repeat data from the GPS and the relative transform from ICP provides an RMSE of 0.208 m in translation and 0.036 rad. 
This localization error is higher than previous works \cite{Burnett2022} but is small enough for the rover to track the path accurately. 

\begin{figure}
    \centering
    \includegraphics[width=0.49\textwidth]{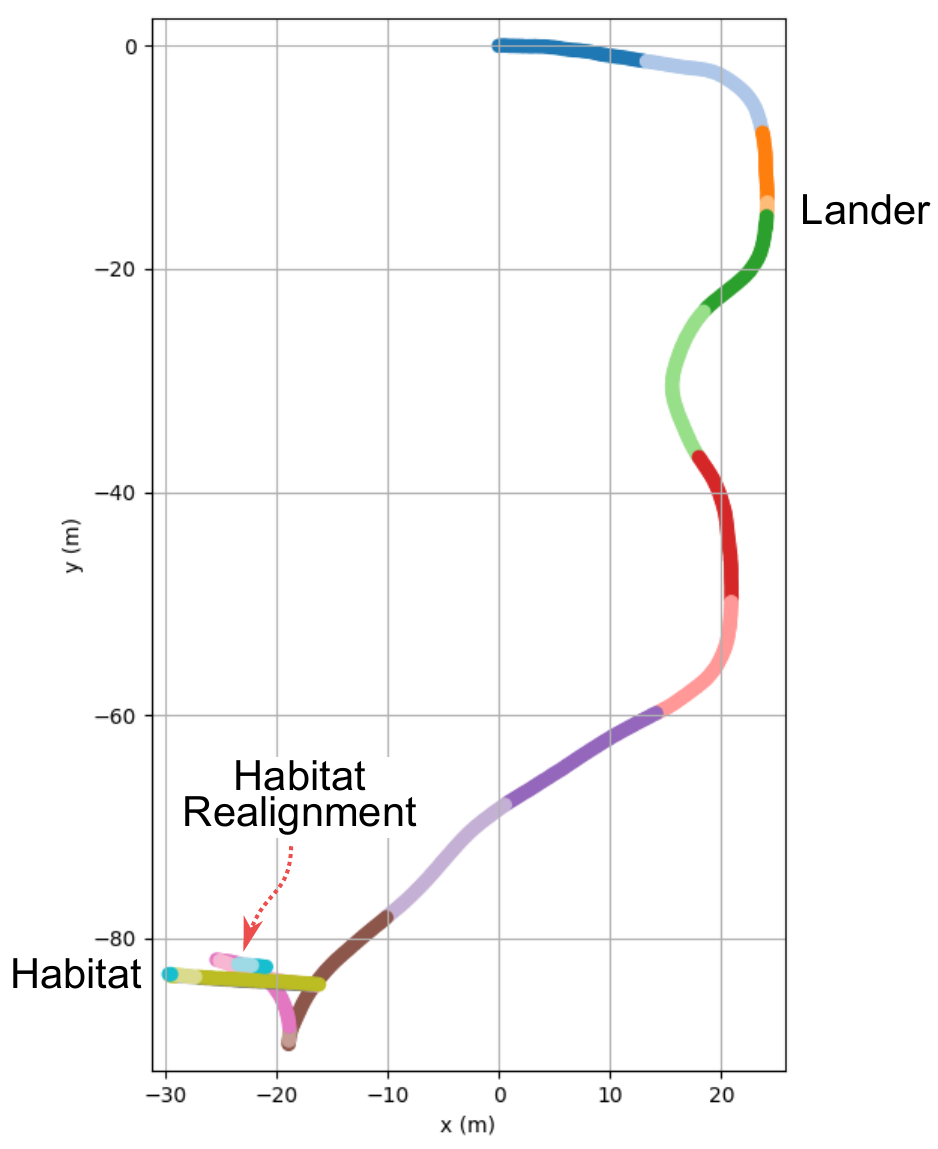}
    \caption{The 18 path segments executed during the semi-autonomous driving portion of the mission. These segments were recorded to create the smooth teach path that was repeated on afterwards. Note that segments are short during precise alignments. }
    \label{fig:path_segments}
\end{figure}

The semi-autonomous teach was built from 18 path segments that were followed by the rover.
In open stretches, the longest path was 15.92 m long. 
In the alignment steps, shortest path was 0.33 m. 
The total path length is 164.2 m long.
These segments are visualized with different colours in \autoref{fig:path_segments}.
There were no e-stop interventions activated. 
The operators' terrain assessment and path execution kept the rover away from hazards. 
Near the habitat, the extra motion occurs due to the vehicle not tracking a maximum steering angle turn precisely. 
This lead the operators to direct the rover to move out and align again. 

The timing breakdown for the mission is compared between semi-autonomous control and autonomous repeating in \autoref{tab:timings}.
In total, driving semi-autonomously took 2:02:14. The same distance of autonomous repeating took 10:32, only 8.6\% of the time.
The cargo lift time could be sped up with a fully automatic procedure. 
Driving out from under the cargo was always performed autonomously. This short repeat took 55 seconds. 

\begin{table}[]
    \centering
    \caption{The duration of mission steps for both forward missions. The speed-up of continuous motion is clear in the precise alignment with the habitat. }
    \begin{tabular}{c|p{2cm}|p{2cm}}
        \textbf{Mission Step} & \textbf{Semi-Autonomous \newline Duration (mm:ss)} &  \textbf{Autonomous \newline Duration (mm:ss)}\\
        \hline
        Drive Start to Lander & 25:20 & 2:42 \\
        Load Cargo & 16:54 & N/A \\
        Drive Lander to Apex & 41:55 & 5:44 \\
        Align from Apex to Habitat & 56:59 & 2:06 \\
        Deliver Cargo & 18:00 & N/A \\
    \end{tabular}
    \label{tab:timings}
\end{table}

\section{DISCUSSION}

Many observations from our field test require further investigation or consideration in future designs. 
The quality of the lidar localization suggests that most of the path-tracking error is a result of the path-tracking controller. 
The most significant model mismatch occurs from the MPC assuming that the rear of the robot moves as a unicycle. 
While this assumption is fairly accurate for shallow curves, the response of the Ackermann vehicle was difficult to model accurately. 
Additionally, the tuning of the controller occurred with different inertial parameters. 
The cargo was not yet ready for use, so 180 kg of cylindrical weights were added to the rear deck.
The cargo has a much larger inertia because of the larger volume. 
While the paths are not the same, LT\&R path-tracking error estimates from the earlier tests were 0.06 m RMSE which is significantly smaller than the 0.10 m observed during this demo. 
Future controllers should consider whether or not the cargo is loaded and adapt accordingly. 
Wang et al. \cite{wang_mpc_2024} employ adaptive MPC control for an autonomous forklift that has dynamics that change with the type of cargo moved.
The low-level torque controller only considered payloads weighing up to 80 kg \cite{chhabra_dynamical_2016} which will compound the problem.

Another issue with the unicycle assumption is the implicit symmetry of forward and reverse motion. 
Ackermann vehicles in reverse are non-minimum phase dynamically and much more difficult to control. 
Using this vehicle meant that the most precise part of the mission must occur with \lelr \  driving in a mode that is more difficult to control. 
Operationally, we mitigated this by providing a long straight route to the habitat which gave the controller more time to reduce the lateral error. 
This extra direction switch occurs between the apex and Habitat in \autoref{fig:gpsRoute}.
A symmetrical vehicle would be easier to control but considering a bicycle model that accounts for reverse Ackermann control issues should mitigate the issue in future deployments. 

Transferring the cargo from the lander to the vehicle using a crane was challenging. 
The load was supported by one strap which left it free to swing and twist with respect to the crane. 
Strong winds during the final demo exacerbated the cargo motion, but even on earlier test days that were relatively calm, the telehandler alone could prevent the cargo from swinging out of position and colliding with the vehicle's cab. 
The pendulum motion of hanging cargo will be significantly more pronounced on the lunar surface with gravity 1/6th of the Earth's. 
Cargo transfer was not the focus of this field test but lunar solutions that involve cranes will require significant support to constrain the motion of large, heavy items. 
Gill et al. \cite{gill_lunar_2022} investigate a number of cargo offloading concepts. 
They note that an important consideration for cargo offloading is whether landers would integrate a payload-specific offloading solution or a general-purpose solution, like the mobile crane used here, would be more cost-effective. 
There is a trade-off between heavier cargo that are self-contained and lighter cargo that require extra infrastructure. 
Observations of this field deployment suggest that alternatives to a mobile crane should be considered. 

At both locations where precise alignment was required, further automation would improve the accuracy and decrease time. 
Automatic alignment would allow more flexibility for landers in various positions.
The time it took to make small, manually computed adjustments to the rover's position near the lander and habitat was significantly longer than the average rate of driving. 
This occurred for a few reasons: first, due to the non-holonomic constraints of the vehicle, correcting lateral errors required backing out a significant distance and trying again; 
second, sensing blind spots made it more difficult to ascertain exactly where the parts at risk of a collision were located; and third, the error margin of path completion was not tunable.
While aligning with the habitat, the drive team decided that the lateral error was greater than 10 cm and out of spec on the first approach. 
To correct this, a new alignment approach would be performed. 
Because every motion was being taught, the drive team requested that the robot repeat back along the path 5 m from the habitat.
Once the robot had completed the repeat, a new teach branch was created with the second successful alignment. 

Part of the challenge with alignment was the occlusions of useful sensors. 
At the lander, the rear lander leg was blocked in the OS-1 lidar scan by the roof lights on the cab, which made it more difficult to assess whether or not the vehicle was parallel. 
Once the cargo was loaded, the lander was completely blocked from lidar scanning. 
At the habitat, the rear camera and lidar gave a clear view of the reverse approach.
However, the cargo should be instrumented with a distance sensor at the docking location.
It was unreliable to assess the longitudinal position of the cargo on \lelr \ once it was loaded.
This created uncertainty of $\pm 7.5$ cm on the location of the cargo. 
While reversing, a conservative envelope had to be used even though the cargo was ultimately loaded almost at the center.
This resulted in being 17 cm away from the habitat rather than less than 15 cm as was the target. 
A direct measurement of the alignment parameters would improve teleoperations and allow for a fully automatic approach to be developed. 

The last capability to improve alignment is a tuneable speed profile for the rover. 
The current path-tracker has the same target and limit speeds independent of the path length. 
For short motions less than 30 cm, the behaviour needs to change.
First, the rover should move more slowly to prevent damage in case the alignment is suboptimal.
Second, the tolerance to decide whether the end of the path has been reached should decrease. 
For long motions, it was satisfactory to use $\pm$5 cm as a complete path threshold, but for a motion that might only be 10 cm, this did not work. 
The drive team added offsets to the desired motion to account for the lower-level behaviour, which should be avoided. 

The wireless communication between the rover and the ground station was mostly reliable. 
However, there were occasional signal dropouts.
The omnidirectional antenna used at the ground station has two stronger lobes. 
Aligning one stronger lobe with the most common location of the rover improved the stability. 
Longer autonomous repeats were not affected directly by communication dropouts because no operator interaction was required. 
One future extension of using LT\&R while exploring would be a communication recovery behaviour where the rover could backtrack along the path to a location with a reliable communication link.

\section{LESSONS LEARNED}

Performing a two-week deployment provided valuable insights into the mechanical system design, software user interfaces, and operational procedures. 

Performing field testing in the Canadian winter meant that snow was anticipated but the weather variability was underappreciated in project planning.
Throughout the deployment, the weather proceeded from dry and cold, to 10 cm of snowfall, to strong winds, to 24 hrs of rain, which melted the snow and to scattered ice patches when the temperature dropped below freezing again. 
The external sensors were IP-rated and could withstand snowfall but the weather pushed the system beyond some of its original design specifications. 
Traction became a variable that was unaccounted for. 
When the sand was dry, the rubber tires left clear tread marks and displaced the sand while driving. 
When the snow fell, the traction was reduced, but with packing snow the rover still drove well. 
But when the wet ground froze over, the rover had trouble navigating as reliably as before. 
The ground was frozen, and the tires left almost no marks in the sand as the rover moved. 
Additionally, localized wheel slip was visible when watching the rover. 
The rover never lost traction overall but the low-level controller was less able to produce the desired body motions. 

Another challenge was site safety conditions for safety operators walking around the test site. 
Snow clearing and salting efforts were taken to prepare the site but the amount of human labour required was not anticipated. 
The rover safety officer handling the emergency stop ensures that the rover and test site are not damaged if something goes wrong. 
When the temperature is below freezing and windy, proper warming gear is critical to allow testing to proceed. 
Large-button e-stops, remote controls, and walkie-talkies had to be implemented so that operators would not have to remove their gloves to activate the robot.
While it might seem convenient to schedule field deployments for summer, it is useful to challenge the system's robustness even if these weather variations are not directly applicable on the Moon. 
Additionally, the earlier sunset during the winter provides more testing time under lunar illumination.

One interesting result was the robustness of the lidar localization to snow accumulation.
Teach passes were completed to build maps before the snow fell. 
\LELR \ could easily localize its pose and then follow the original path even though the snow obscured many small geometric features such as rocks. 
With the amount of appearance change, a human operator could not have followed the original path.

Using teach and repeat as an operator motivates several useful back-end changes to better support cargo missions. 
Previous deployments of teach and repeat do not contain tight alignment stages in the middle of a route.
Even when driving with the joystick, an alignment attempt would be made while teaching but the final position or path was deemed unsatisfactory. 
If a teach is continued while backing up and driving forward to align again, all subsequent repeats would include the erroneous alignment and direction switch.
In this field trial, we did a short repeat back to an upstream location along the path, branched onto a new teach and tried the alignment again. 
This achieved the desired effect as future repeats all ignore the original stub branch.
However, this created an operational problem as the two routes would be so close to each other that the visualization would often overlap. 
Without a layer selection, it was difficult to ensure that a target waypoint was on the correct branch. 
This motivates a new feature to mark branches as stale so they do not appear in the user interface as locations for waypoint selection. 

Another useful feature during alignment would be a ``bookmark position'' button that forces a pose-graph vertex to be exactly in the current rover location. 
To reduce storage requirements and smooth high-frequency oscillations caused by odometry, LT\&R only stores a pose every 30 cm or 10$^\circ$. 
The ideal aligned location is unlikely to sit exactly 30 cm away, especially in a place such as the lander, which is not at the end of the path. 
A workaround was developed by stopping and restarting the teach in the target location. 
This is undesirable because the two teach segments are no longer locally consistent through an odometry rollout and lead to tiny direction switches or jumps in the target path. 

The final operator feature to add is an offset from the waypoint for the end of paths. 
When \lelr \ aligned autonomously to the habitat, the teach took the cargo until it was less than 5 cm from the contact interface.
While LT\&R could achieve this precision, out of caution there was interest to make the initial alignment have a gap and then proceed. The smallest gap that could be requested was 30 cm which did not bring the rover acceptably close. 
Adding the functionality to go to the end minus 10 cm would have been ideal.

\section{CONCLUSION}
Through an extensive two-week deployment of \ifx\undefined\blind the Lunar Exploration Light Rover, \else a one-tonne path-to-flight rover prototype, \fi a successful cargo transport concept was demonstrated and evaluated. 
The mission concept considered the case of moving cargo between the same lander and habitat repeatedly.
Lidar teach and repeat was demonstrated as an effective framework for precise navigation in a lunar analogue environment. 
Cargo alignment was within 20 cm of contact, path tracking errors were about 25 cm RMSE and repeating took only 8.6\% of the time of semi-autonomous motion.
Future extensions of the concept include automatic alignment of the rover to the habitat and lander.
Additional monitoring of the route may be required to ensure that the taught path remains stable. 
Change detection techniques could be used to monitor the effects of the terrain looking for new obstacles or growing trenches from the rover's repeated motion.
These improvements would improve the resilience of the system to changes in the lunar environment caused by the Artemis program.
This capability would unlock functionality that supports cargo landers who do not land in the same spot every time.

\section{ACKNOWLEDGMENT}
\ifx\blind\undefined
We thank the team at the Canadian Space Agency, David Gingras, Fiona Kirby, Cody Barth, Martin Picard, and Tongxi Wu, for their support with the Analogue Terrain and operations. 
Additionally, we thank Igor Coelho Correia for taking many of the detailed pictures used in this paper.
We thank Hexagon for the use of Inertial Xplorer.
\else
\textbf{Omitted for anonymous review.}
\fi

\bibliographystyle{./bib/IEEEtran}
\bibliography{./bib/bib}

\end{document}

%% file: images/alignment_dual.pdf_tex
\begingroup%
  \makeatletter%
  \providecommand\color[2][]{%
    \errmessage{(Inkscape) Color is used for the text in Inkscape, but the package 'color.sty' is not loaded}%
    \renewcommand\color[2][]{}%
  }%
  \providecommand\transparent[1]{%
    \errmessage{(Inkscape) Transparency is used (non-zero) for the text in Inkscape, but the package 'transparent.sty' is not loaded}%
    \renewcommand\transparent[1]{}%
  }%
  \providecommand\rotatebox[2]{#2}%
  \newcommand*\fsize{\dimexpr\f@size pt\relax}%
  \newcommand*\lineheight[1]{\fontsize{\fsize}{#1\fsize}\selectfont}%
  \ifx\svgwidth\undefined%
    \setlength{\unitlength}{632.61968994bp}%
    \ifx\svgscale\undefined%
      \relax%
    \else%
      \setlength{\unitlength}{\unitlength * \real{\svgscale}}%
    \fi%
  \else%
    \setlength{\unitlength}{\svgwidth}%
  \fi%
  \global\let\svgwidth\undefined%
  \global\let\svgscale\undefined%
  \makeatother%
  \begin{picture}(1,0.33647868)%
    \lineheight{1}%
    \setlength\tabcolsep{0pt}%
    \put(0.175,0.01948887){\color[rgb]{0,0,0}\makebox(0,0)[lt]{\lineheight{1.25}\smash{\begin{tabular}[t]{l}(a) Lander Alignment\end{tabular}}}}%
    \put(0.65,0.02166274){\color[rgb]{0,0,0}\makebox(0,0)[lt]{\lineheight{1.25}\smash{\begin{tabular}[t]{l}(b) Habitat Alignment\end{tabular}}}}%
    \put(0,0){\includegraphics[width=\unitlength,page=1]{./images/alignment_dual.pdf}}%
  \end{picture}%
\endgroup%

%% file: images/main_figure_small_blurred.pdf_tex
\begingroup%
  \makeatletter%
  \providecommand\color[2][]{%
    \errmessage{(Inkscape) Color is used for the text in Inkscape, but the package 'color.sty' is not loaded}%
    \renewcommand\color[2][]{}%
  }%
  \providecommand\transparent[1]{%
    \errmessage{(Inkscape) Transparency is used (non-zero) for the text in Inkscape, but the package 'transparent.sty' is not loaded}%
    \renewcommand\transparent[1]{}%
  }%
  \providecommand\rotatebox[2]{#2}%
  \newcommand*\fsize{\dimexpr\f@size pt\relax}%
  \newcommand*\lineheight[1]{\fontsize{\fsize}{#1\fsize}\selectfont}%
  \ifx\svgwidth\undefined%
    \setlength{\unitlength}{460.68901062bp}%
    \ifx\svgscale\undefined%
      \relax%
    \else%
      \setlength{\unitlength}{\unitlength * \real{\svgscale}}%
    \fi%
  \else%
    \setlength{\unitlength}{\svgwidth}%
  \fi%
  \global\let\svgwidth\undefined%
  \global\let\svgscale\undefined%
  \makeatother%
  \begin{picture}(1,1.33352555)%
    \lineheight{1}%
    \setlength\tabcolsep{0pt}%
    \put(0.15507205,1.0266968){\color[rgb]{0,0,0}\makebox(0,0)[lt]{\lineheight{1.25}\smash{\begin{tabular}[t]{l}(a) Rover at the Start\end{tabular}}}}%
    \put(0.63996532,1.02686796){\color[rgb]{0,0,0}\makebox(0,0)[lt]{\lineheight{1.25}\smash{\begin{tabular}[t]{l}(b) Rover aligned with the Lander\end{tabular}}}}%
    \put(0.10410493,0.6859349){\color[rgb]{0,0,0}\makebox(0,0)[lt]{\lineheight{1.25}\smash{\begin{tabular}[t]{l}(c) Rover rounding the South curve\end{tabular}}}}%
    \put(0.68483727,0.68592745){\color[rgb]{0,0,0}\makebox(0,0)[lt]{\lineheight{1.25}\smash{\begin{tabular}[t]{l}(d) Rover at the Apex\end{tabular}}}}%
    \put(0.10779505,0.34486035){\color[rgb]{0,0,0}\makebox(0,0)[lt]{\lineheight{1.25}\smash{\begin{tabular}[t]{l}(e) Cargo aligned with the Habitat\end{tabular}}}}%
    \put(0.67720514,0.34477837){\color[rgb]{0,0,0}\makebox(0,0)[lt]{\lineheight{1.25}\smash{\begin{tabular}[t]{l}(f) Cargo legs extended\end{tabular}}}}%
    \put(0.13128163,0.00378589){\color[rgb]{0,0,0}\makebox(0,0)[lt]{\lineheight{1.25}\smash{\begin{tabular}[t]{l}(g) Cargo lifted off the rover\end{tabular}}}}%
    \put(0.62628147,0.00378569){\color[rgb]{0,0,0}\makebox(0,0)[lt]{\lineheight{1.25}\smash{\begin{tabular}[t]{l}(h) Rover drives away from the cargo\end{tabular}}}}%
    \put(0,0){\includegraphics[width=\unitlength,page=1]{./images/main_figure_small_blurred.pdf}}%
    \put(0.17671799,1.24138085){\color[rgb]{1,1,1}\makebox(0,0)[lt]{\lineheight{1.25}\smash{\begin{tabular}[t]{l}Rover\end{tabular}}}}%
    \put(0,0){\includegraphics[width=\unitlength,page=2]{./images/main_figure_small_blurred.pdf}}%
  \end{picture}%
\endgroup%